\pdfoutput=1

\documentclass[11pt]{article}

\usepackage{acl}

\usepackage{times}
\usepackage{latexsym}

\usepackage[T1]{fontenc}

\usepackage[utf8]{inputenc}

\usepackage{microtype}

\usepackage{inconsolata}

\usepackage{graphicx}

\usepackage{booktabs}
\usepackage{arydshln}
\usepackage{multirow}
\usepackage{amsmath}

\definecolor{mlp_blue}{HTML}{1f77b4}
\definecolor{mlp_orange}{HTML}{ff7f0e}
\definecolor{mlp_green}{HTML}{2ca02c}

\title{Out-of-the-Box Conditional Text Embeddings \\ from Large Language Models}

\author{Kosuke Yamada \and Peinan Zhang \\
Cyberagent Inc., Japan \\
\texttt{\{yamada\_kosuke,zhang\_peinan\}@cyberagent.co.jp}}

\begin{document}
\maketitle
\begin{abstract}
Conditional text embedding is a proposed representation that captures the shift in perspective on texts when conditioned on a specific aspect.
Previous methods have relied on extensive training data for fine-tuning models, leading to challenges in terms of labor and resource costs.
We propose \textbf{PonTE}, a novel unsupervised conditional text embedding method that leverages a causal large language model and a conditional prompt.
Through experiments on conditional semantic text similarity and text clustering, we demonstrate that PonTE can generate useful conditional text embeddings and achieve performance comparable to supervised methods without fine-tuning.
We also show the interpretability of text embeddings with PonTE by analyzing word generation following prompts and embedding visualization.
\end{abstract}

\section{Introduction}
\label{sec:introduction}
Text embeddings, which represent text as dense semantic vectors, are essential for various natural language processing (NLP) tasks, such as semantic textual similarity~\cite{agirre-etal-2012-semeval,marelli-etal-2014-sick,cer-etal-2017-semeval} and text clustering tasks~\cite{aggarwal2012survey,wehrli-etal-2023-german}.
Many text embedding models~\cite{conneau-etal-2017-supervised,cer-etal-2018-universal,reimers-gurevych-2019-sentence,gao-etal-2021-simcse} aim to generate a single universal representation for a given text. 
However, this universal representation faces challenges when addressing the inherent ambiguity in text.
For example, consider the review texts shown in Table~\ref{tab:examples}.
$T_1$ and $T_2$ discuss similar product categories but offer different evaluations, whereas $T_1$ and $T_3$ discuss different product categories but provide similar evaluations.
When measuring the similarity between them without constraints, it becomes challenging to determine whether the similarity between $T_1$ and $T_2$ or between $T_1$ and $T_3$ should be higher.

\begin{table}[!t]
\vspace{-5pt}
\small
\centering
\begin{tabular}{l} 
\toprule
\textcolor{mlp_blue}{$T_1$}: This camera is one of my favorites. \\
\textcolor{mlp_orange}{$T_2$}: This smartphone cannot capture high-quality images. \\
\textcolor{mlp_green}{$T_3$}: Best fish I have ever had. \\ 
\bottomrule
\end{tabular}
\vspace{-5pt}
\caption{Examples of semantically ambiguous texts}
\label{tab:examples}
\end{table}

\begin{figure}[!t]
\centering
\vspace{-5pt}
\includegraphics[width=\linewidth]{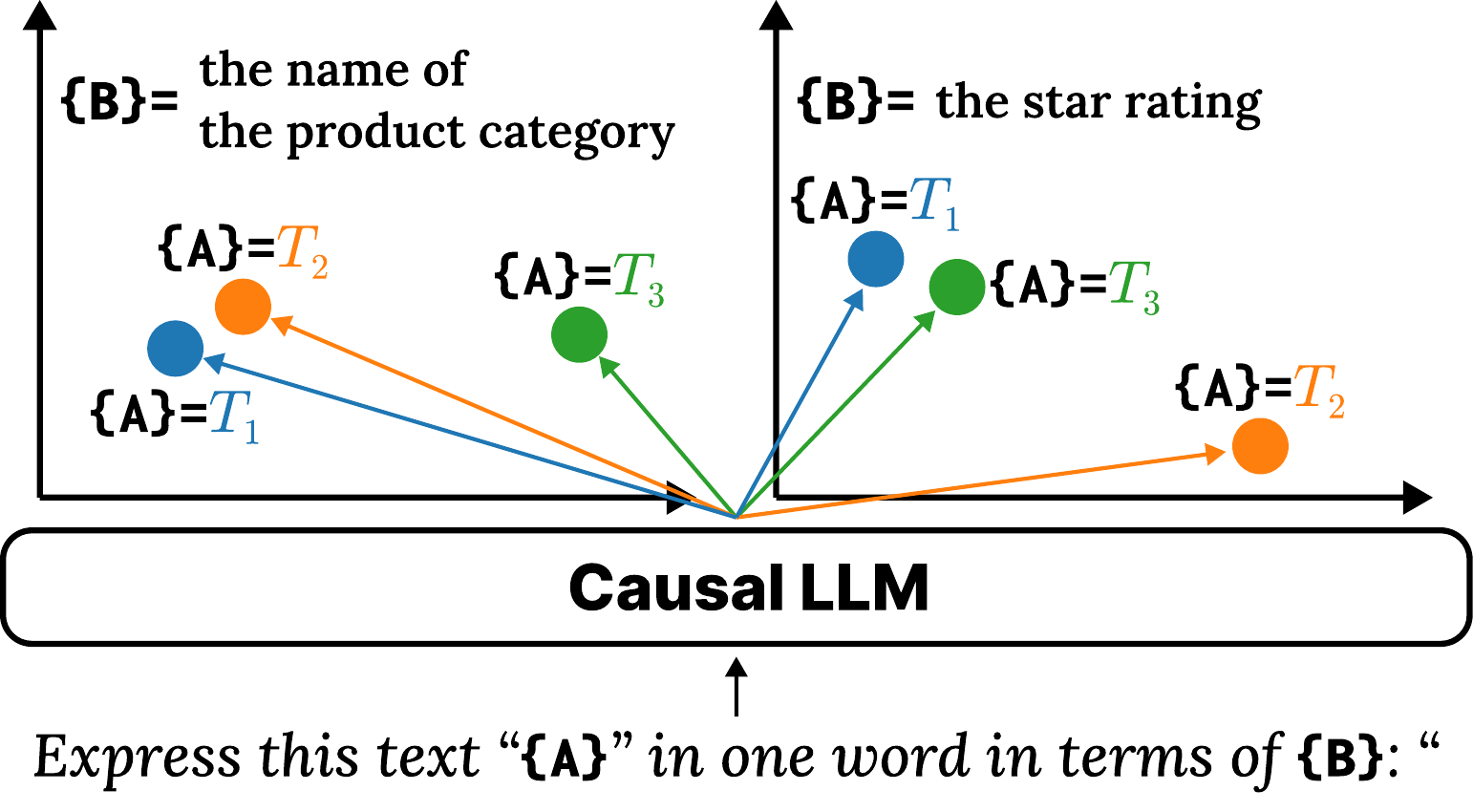}
\vspace{-20pt}
\caption{Visualization examples of conditional text embedding with PonTE. 
\textcolor{mlp_blue}{$T_1$}, \textcolor{mlp_orange}{$T_2$}, and \textcolor{mlp_green}{$T_3$} correspond to the texts in Table~\ref{tab:examples}.}
\label{fig:overview}
\vspace{-10pt}
\end{figure}

Given this context, conditional text embeddings, where texts are embedded by specifying aspects as conditions, have been proposed~\cite{deshpande-etal-2023-c,yoo2024hyper}.
However, previous methods require training data for a conditional semantic text similarity (C-STS) task and are not easily applicable to NLP tasks beyond the domain and the language.
Also, instruction fine-tuning embedding methods are a type of conditional text embedding method in that they generate different embeddings for each task~\cite{su-etal-2023-one,li2023generaltextembeddingsmultistage,wang2024improvingtextembeddingslarge} since these methods generate task-specific text embeddings by inputting prompts consisting of an instruction and text into a causal large language model (LLM).
However, these methods require fine-tuning the LLMs using a significant amount of manually or automatically generated training data, which is laborious, time-consuming, uneconomical, and computationally burdensome.

In this study, we propose \textbf{PonTE}, a \textbf{P}rompt-based c\textbf{on}ditional \textbf{T}ext \textbf{E}mbedding method, as shown in Figure~\ref{fig:overview}.
PonTE is a method that uses causal LLMs as text embedders with conditional prompts inspired by PromptEOL~\cite{jiang2023scaling}.
In PonTE, we input the prompt `\textit{Express this text ``{\rm \{A\}}'' in one word in terms of {\rm \{B\}}: ``}', where \{A\} is the target text and \{B\} is the conditional text, into the LLM and extract the hidden state vector in the model as the conditional text embedding.
PonTE enables the generation of effective conditional text embeddings by condition-oriented compression of textual information.

\section{PonTE}
\label{sec:methodology}
In the conditional text embedding, there are only supervised methods that use domain data or need to be fine-tuned on large-scale datasets, and domain- and language-agnostic versatile unsupervised methods have not yet been fully explored.
To develop a general-purpose method without costly effort, we propose PonTE, which is a prompt-based conditional text embedding method.

\begin{figure}[t]
\centering
\includegraphics[width=\linewidth]{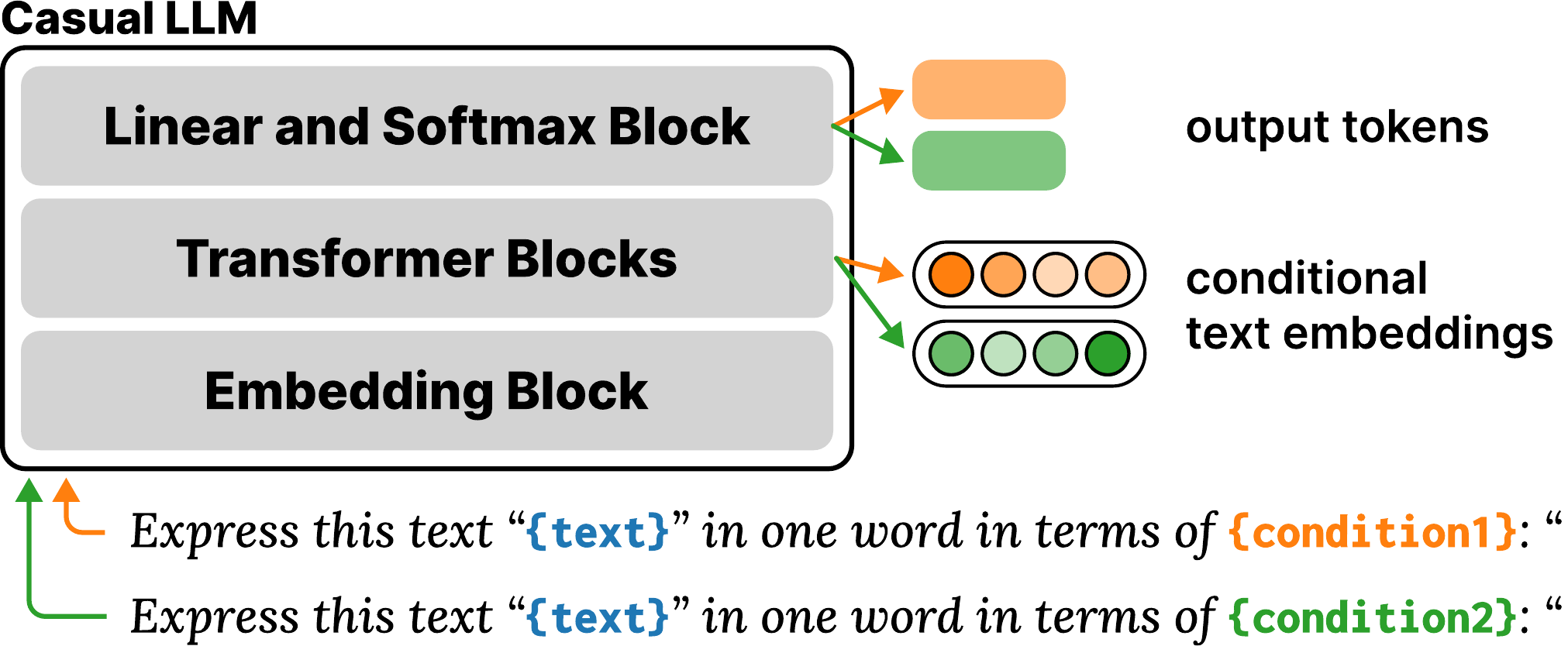}
\vspace{-20pt}
\caption{Overview of PonTE}
\label{fig:method}
\vspace{-15pt}
\end{figure}

An overview of PonTE is shown in Figure~\ref{fig:method}.
We utilize the hidden state vector within the transformer blocks of the causal LLM of the last token in the conditional prompt as the conditional text embedding.
PonTE introduces one-word and conditional restrictions in the prompts.
For example, we employ the prompt templates such as `\textit{This text: ``{\rm \{text\}}'' means in one word in terms of {\rm \{condition\}}: ``}', which is a similar one used in PromptEOL, or `\textit{Express this text ``{\rm \{text\}}'' in one word in terms of {\rm \{condition\}}: ``}', which is aware of the instruction-tuned model.
Within the prompts, the target text is input in `\{text\}', whereas the conditional text is input in `\{condition\}'.
By forcing the causal LLM to represent text in single word while specifying the aspect as the condition, it is expected to compress the textual information in a purposeful manner.

We can also interpret the embedding with PonTE by generating a single word that follows the prompt through the linear and softmax block.
This word refers to the concatenation of output tokens until `\textit{''}' is generated and leads to a better understanding of the prediction results and helps to determine prompts and analyze errors.

\section{Experiments on C-STS}
\label{sec:csts}
We experiment with a C-STS task that measures the semantic similarity between texts based on given conditions to confirm PonTE's capability in generating text embeddings based on specified conditions.

\subsection{Settings}
\label{subsec:csts_settings}
We used the C-STS dataset~\cite{deshpande-etal-2023-c}.\footnote{Statistics on the datasets are presented in Appendix~\ref{sec:dataset}.}
Each instance comprises a quadruple consisting of two texts, a condition, and a gold similarity.
For the evaluation, Spearman's rank correlation coefficient~($\rho$) and Pearson's correlation coefficient~($r$) are used based on a predicted similarity and the gold similarity of the text pairs.
The predicted similarity is computed by calculating the cosine similarity between text embeddings obtained by inputting conditional prompts into the models.

As causal LLMs for PonTE, we used the base and instruction-tuned Mistral-7B~\cite{jiang2023mistral7b}, the base and instruction-tuned Llama-3-8B~\cite{touvron2023llama}, and the base and instruction-tuned Llama-3-70B as causal LLMs.\footnote{The model links are in Appendix~\ref{sec:specific_models}.}
The prompt templates are detailed in Appendix~\ref{sec:csts_prompt_template_search}.

We conducted a comparison between PonTE and two instruction fine-tuned embedding methods trained on diverse and massive datasets for causal LLMs, which have demonstrated high performance in recent MTEB~\cite{muennighoff-etal-2023-mteb}.
The methods were GTE~\cite{li2023generaltextembeddingsmultistage}, which is based on the instruction-tuned Qwen2-7B, and E5~\cite{wang2024textembeddingsweaklysupervisedcontrastive,wang2024improvingtextembeddingslarge}, which is based on the instruction-tuned Mistral-7B. 
Additionally, we compared PonTE with the bi-encoder method with unsupervised and supervised SimCSE.\footnote{An explanation of these methods is in Appendix \ref{sec:additional_explanation_previous_methods}.}

\subsection{Results}
\label{subsec:csts_results}
The experimental results are listed in Table~\ref{tab:experiments_csts}.
PonTE consistently demonstrated high performance across different models.
In particular, PonTE$_{\textrm{\footnotesize Llama-3-8B-Inst}}$ achieved high scores with Spearman and Pearson correlation coefficients of 37.1 and 33.6, respectively.
Even without fine-tuning, the proposed method achieves scores comparable to state-of-the-art supervised methods.

\begin{table}[t!]
\vspace{-5pt}
\small
\centering
\begin{tabular}{l cc} \\ 
\toprule
Method & $\rho$ & $r$ \\ 
\midrule \midrule
\multicolumn{3}{c}{\textit{Supervised method}} \\ 
\midrule
sup-SimCSE$_{\textrm{\footnotesize large}}$ & 3.4 & 4.1 \\ 
\midrule
GTE$_{\textrm{\footnotesize Qwen2-7B-Inst}}$ & 33.5 & 33.9 \\
E5$_{\textrm{\footnotesize Mistral-7B-Inst}}$ & \textbf{34.8} & \textbf{34.6} \\
\midrule \midrule

\multicolumn{3}{c}{\textit{Unsupervised method}} \\ 
\midrule
unsup-SimCSE$_{\textrm{\footnotesize large}}$ & 2.3 & 1.7 \\ 
\midrule
PonTE$_{\textrm{\footnotesize Mistral-7B}}$ & 21.6 & 21.0 \\
PonTE$_{\textrm{\footnotesize Mistral-7B-Inst}}$ & 30.6 & 28.9 \\
PonTE$_{\textrm{\footnotesize Llama-3-8B}}$ & 21.7 & 19.7 \\
PonTE$_{\textrm{\footnotesize Llama-3-8B-Inst}}$ & \textbf{37.1} & \textbf{33.6} \\ 
PonTE$_{\textrm{\footnotesize Llama-3-70B}}$ & 11.3 & 10.9 \\
PonTE$_{\textrm{\footnotesize Llama-3-70B-Inst}}$ & 35.1 & 31.0 \\ 
\bottomrule
\end{tabular}
\vspace{-5pt}
\caption{Experimental results on C-STS. 
\textbf{Bold} indicates the highest score in each setting.}
\label{tab:experiments_csts}
\vspace{-15pt}
\end{table}

\begin{table*}
\small
\centering
\begin{tabular}{@{\ \ }l@{\ \ }l@{\ \ }ll@{\ \ }c@{\ \ }c@{\ \ }l@{\ \ }l@{\ }} \\ 
\toprule
& Text1 & Text2 & Condition & Gold & Pred & Word1 & Word2 \\
\midrule

\multirow{4}{*}{(a)}
& $T_{a1}$: A group of elephants of
& $T_{a2}$: One elephant is squirting 
& $C_{a1}$: the phy-
& \multirow{2}{*}{1.0} & \multirow{2}{*}{1.30}
& \multirow{2}{*}{Walking} & water- \\ 

& different sizes walking together 
& water out of its mouth 
& ical actions
&&&
& squirting \\ \cline{4-8}

& on dirt with a rock formation 
& and the other is putting
& \multirow{1.5}{*}{$C_{a2}$: the}
& \multirow{2.5}{*}{5.0} & \multirow{2.5}{*}{4.77} 
& \multirow{2.5}{*}{Elephants} & \multirow{2.5}{*}{Elephant} \\ 

& and trees in the background.
& water into its mouth.
& \multirow{1.5}{*}{animal} 
&&&& \\ 
\midrule

\multirow{4}{*}{(b)}
& $T_{b1}$: A man in a shirt and tie 
& $T_{b2}$: The man is wearing a 
& $C_{b1}$: the attire 
& \multirow{2}{*}{2.0} & \multirow{2}{*}{4.52}
& \multirow{2}{*}{Formal} & \multirow{2}{*}{Formal} \\

& with his hands in his pockets 
& dress coat, suit and tie, 
& of the person
&&&& \\ \cline{4-8}

& leaning against a wall.
& but not dress pants.
& \multirow{1.5}{*}{$C_{b2}$: the gender}
& \multirow{2.5}{*}{5.0} & \multirow{2.5}{*}{4.64}
& \multirow{2.5}{*}{Male} & \multirow{2.5}{*}{Male} \\

&&& \multirow{1.5}{*}{of the person}
&&&& \\ 
\bottomrule
\end{tabular}
\vspace{-5pt}
\caption{Output examples for PonTE with Llama-3-8b-Inst.
``Pred'' denotes the min-max scaled value of the cosine similarity by 0.5 and 5.5.
``Word1'' and ``Word2'' denote the generated words at ``Text1'' and ``Text2,'' respectively.}
\label{tab:examples_csts}
\end{table*}

A comparison between E5$_{\textrm{\footnotesize Mistral-7B-Inst}}$ and PonTE$_{\textrm{\footnotesize Mistral-7B-Inst}}$ confirms that the fine-tuning for the text embedding model is valid, as E5$_{\textrm{\footnotesize Mistral-7B-Inst}}$ performs better.
If E5$_{\textrm{\footnotesize Mistral-7B-Inst}}$ were constructed, it could potentially achieve higher performance.
However, E5$_{\textrm{\footnotesize Mistral-7B-Inst}}$ is trained on 500k examples generated using 150k unique instructions and a total of 1.8 million manually generated examples in the QA and search datasets, which is not easy to build and expensive to develop.
Also, the performance of SimCSE is poor because it cannot interact between target and conditional texts without fine-tuning using manual data.
We suggest that PonTE could serve as a new baseline for unsupervised conditional text embedding, which has not been extensively explored.

We validated that instruction-tuned models outperform non-instruction-tuned models under PonTE.
This could be attributed to the fact that instruction-tuning facilitates adherence to the instruction in the conditional prompt.
In addition, the comparison of PonTE based on Llama-3-8B and Llama-3-70B showed no difference in performance.
According to \citet{jiang2023scaling}, scaling up does not directly correlate with performance in their experiment on STS for PromptEOL using OPT with varying numbers of parameters without fine-tuning.
Our results are consistent with theirs.

\subsection{Analysis}
The output examples for PonTE with Llama-3-8B-Inst are listed in Table \ref{tab:examples_csts}.\footnote{In the C-STS dataset, the test set is not disclosed; therefore, we analyze the results of the validation set.}
It contains two text pairs, two conditions, gold scores and predicted scores for the text pairs, and one generated word for each text.
From Table~\ref{tab:examples_csts}~(a), we can see that PonTE is able to obtain appropriate prediction scores and generated words according to the conditions for each text.
From the results, we obtained new insights of the deep relationship between prediction scores and generated words.
Table~\ref{tab:examples_csts}~(b) shows that PonTE predicts the same word for each text, despite the low gold similarity, and has an inappropriately high prediction score.
This issue is due to the complexity of the task of summarizing a text in a single word.

\begin{figure}[!t]
\vspace{-10pt}
\centering
\includegraphics[width=0.98\linewidth]{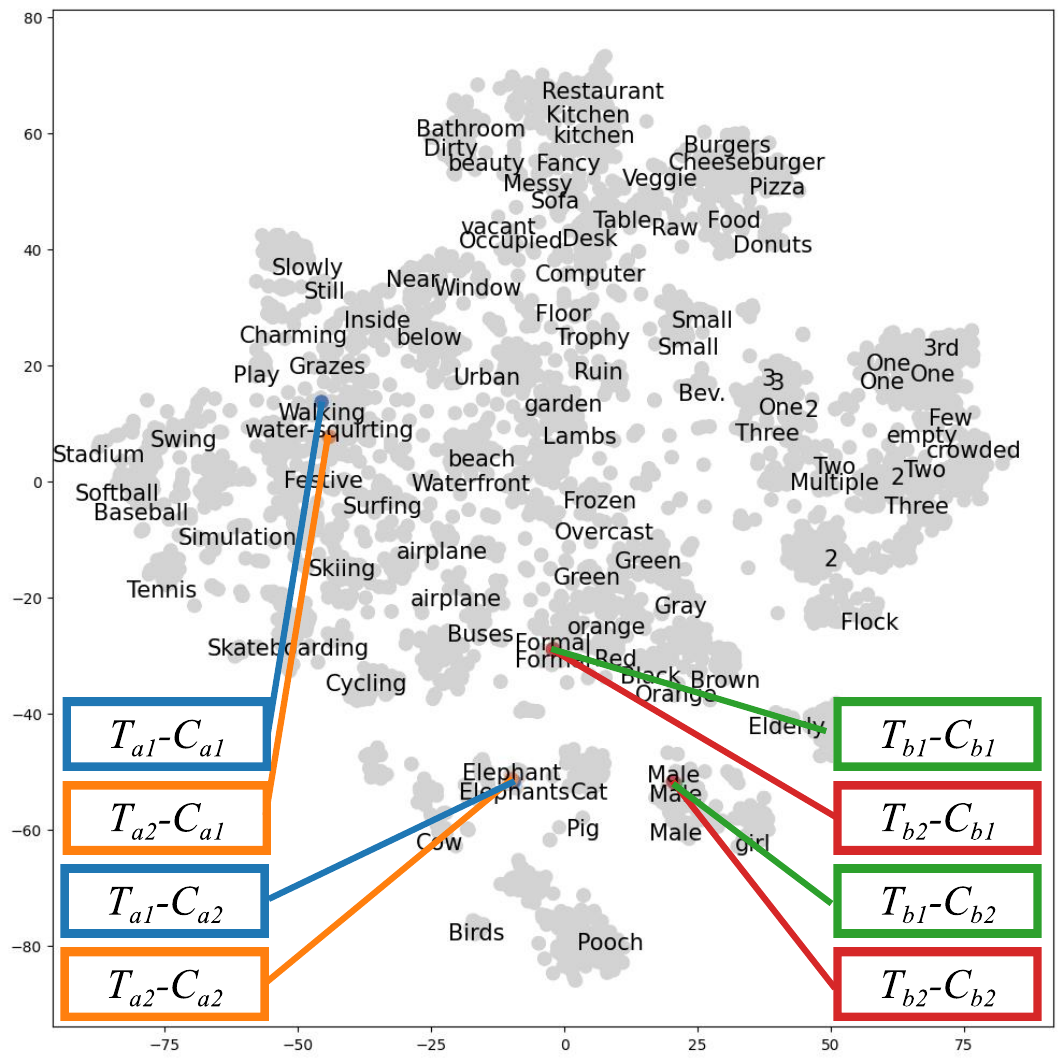}
\vspace{-5pt}
\caption{2D t-SNE projection of embedding by PonTE}
\label{fig:visualization_csts}
\vspace{-10pt}
\end{figure}

The visualization results of text embeddings by PonTE$_{\textrm{\footnotesize Llama-3-8B-Inst}}$ projected into two dimensions by t-SNE~\cite{van2008visualizing} are shown in Figure~\ref{fig:visualization_csts}.
Each point represents the embedding for each text and displays the one generated word.
The instances (a) and (b) from Table~\ref{tab:examples_csts} are highlighted.
The findings reveal that identical texts are embedded far apart under each condition, indicating that the texts are embedded based on conditions rather than surface information.
Also, the similar words are clustered closely together in the semantic space. 
For example, numbers are clustered on the right, whereas creatures are clustered at the bottom.
This is new evidence that PonTE is able to project the text into the semantic space of the word.

\section{Experiments on Text Clustering}
We experienced with text clustering, which groups text by its meaning, as one of the promising applications of conditional text embedding.

\subsection{Settings}
Conditional text embedding allows text to be represented from the user's desired aspect.
To assess its feasibility, we performed text clustering from multiple aspects, using the English dataset from Amazon reviews corpus, where each review text was assigned multiple different labels related to category (Amazon-C) and rating (Amazon-R).
We also addressed common text clustering tasks through two datasets: ScienceQA (SciQA)~\cite{lu2022learn}, and Tweet emotion intensity dataset (Tweet Emotion)~\cite{mohammad-bravo-marquez-2017-wassa}.
Amazon-C and SciQA focused on clustering based on topics, whereas Amazon-R and Tweet emotion focused on clustering based on emotions.

The K-means~\cite{macqueen1967some} algorithm is used for clustering, and the number of clusters is given.
The V-measure~\cite{rosenberg-hirschberg-2007-v} served as the evaluation metric, which is independent of the permutations of clustering labels.
Clustering is carried out five times with different seeds, and the average is used as the score.

In PonTE, we utilized models based on Mistral-7B and Llama-3-8B.
We also used the prompt template that gave the best Spearman correlation coefficient in C-STS and the conditional text inserted into the prompt template was selected based on its V-measure score in the validation set.\footnote{Details are provided in Appendix~\ref{sec:csts_validation_set}, ~\ref{sec:clustering_condition_search}, and ~\ref{sec:text_clustering_validation}}
As comparison methods, we incorporated GTE, and E5 utilized in the C-STS experiment.
In addition, we compared PonTE with SimCSE and PromptEOL, which are universal text embedding methods.

\begin{table}[t!]
\vspace{-5pt}
\small
\centering
\begin{tabular}{@{\ \ \ }l@{\ \ \ }c@{\ \ \ }c@{\ \ \ }c@{\ \ \ }c@{\ \ \ }}
\toprule
\multirow{2}{*}{Method} & \multicolumn{2}{@{}c}{Amazon} & Sci- & Tweet \\
& -C & -R & QA & Emo. \\
\midrule \midrule

\multicolumn{5}{c}{\textit{Supervised method}} \\ 
\midrule
sup-SimCSE$_{\textrm{\footnotesize large}}$ & 19.5 & 22.4 & 65.5 & 29.4 \\
\midrule

GTE$_{\textrm{\footnotesize Qwen2-7B-Inst}}$ & \textbf{38.3} & 36.8 & 73.9 & 36.8 \\
E5$_{\textrm{\footnotesize Mistral-7B-Inst}}$ & 37.4 & \textbf{37.6} & \textbf{74.0} & \textbf{41.3} \\

\midrule \midrule

\multicolumn{5}{c}{\textit{Unsupervised method}} \\ 
\midrule

unsup-SimCSE$_{\textrm{\footnotesize large}}$ & 16.7 & 4.2 & 63.8 & 23.4 \\
\midrule

PromptEOL$_{\textrm{\footnotesize Mistral-7B}}$ & 8.6 & 27.2 & 66.0 & 6.5 \\
PromptEOL$_{\textrm{\footnotesize Mistral-7B-Inst}}$ & 6.1 & 27.4 & 59.4 & 22.7 \\
PromptEOL$_{\textrm{\footnotesize Llama-3-8B}}$ & 9.9 & 20.4 & 66.7 & 9.5 \\
PromptEOL$_{\textrm{\footnotesize Llama-3-8B-Inst}}$ & 9.4 & 30.8 & 65.1 & 31.7 \\
\midrule

PonTE$_{\textrm{\footnotesize Mistral-7B}}$ & 27.7 & 27.7 & \textbf{74.5} & 18.1 \\
PonTE$_{\textrm{\footnotesize Mistral-7B-Inst}}$ & 25.3 & 31.7 & 68.0 & 43.8 \\
PonTE$_{\textrm{\footnotesize Llama-3-8B}}$ & \textbf{30.9} & 23.8 & 74.1 & 24.0 \\
PonTE$_{\textrm{\footnotesize Llama-3-8B-Inst}}$ & 30.5 & \textbf{34.1} & 73.0 & \textbf{45.9} \\
\bottomrule
\end{tabular}
\vspace{-5pt}
\caption{Experimental results of text clustering.
The score represents the V-measure value.}
\label{tab:english_text_clustering}
\vspace{-10pt}
\end{table}

\subsection{Results}
The results are listed in Table~\ref{tab:english_text_clustering}.
PonTE outperformed the other unsupervised methods across all datasets and PonTE$_{\textrm{\footnotesize Llama-3-8B-Inst}}$ shows a high and balanced performance.
PonTE outperformed PromptEOL across all models and datasets, highlighting the effectiveness of conditional prompt for each aspect.
Also, PonTE outperformed supervised SimCSE, strong universal text embeddings on all datasets, and on some datasets, PonTE outperformed GTE and E5, which use a variety of training data, including datasets similar to the domain of this task.
In short, PonTE, which can generate useful text embeddings without fine-tuning, is likely to be more useful than GTE and E5 for domains and languages not supported by these models.

\section{Conclusion}
In this study, we proposed PonTE, a novel unsupervised conditional text embedding method that does not require fine-tuning.
PonTE generated high-quality conditional text embeddings by leveraging powerful LLMs and conditional prompts.
Experiments on C-STS and text clustering show that PonTE outperformed existing unsupervised methods and performed comparably to supervised methods.
In the future, we will explore other applications of PonTE and enhance its functionality.

\section*{Limitation}
Our experiments have two limitations.
First, our evaluation tasks were limited to conditional semantic text similarity and text clustering. 
While these tasks were crucial for evaluating the performance of embedding a single text in multiple aspects, evaluating PonTE on a wider range of tasks, such as those included in MTEB~\cite{muennighoff-etal-2023-mteb}, would provide a more comprehensive assessment of its performance.
Furthermore, while our method is language-independent, its performance has not been assessed across different languages.

The second limitation involves prompt engineering, particularly when utilizing LLMs in general.
While fine-tuning is not necessary, validating the performance with additional data is recommended to ensure optimal results.

\bibliography{embeddings}

\appendix

\section{Datasets used in Experiments}
\label{sec:dataset}
In our C-STS experiment, we used the C-STS dataset.\footnote{\href{https://github.com/princeton-nlp/c-sts}{princeton-nlp/c-sts}}
In our text clustering experiment, we used three datasets, Amazon reviews corpus,\footnote{\href{https://www.kaggle.com/datasets/mexwell/amazon-reviews-multi}{mexwell/amazon-reviews-multi}} ScienceQA,\footnote{\href{https://huggingface.co/datasets/derek-thomas/ScienceQA}{derek-thomas/ScienceQA}} and Tweet emotion intensity dataset.\footnote{\href{https://huggingface.co/datasets/cardiffnlp/tweet_eval}{cardiffnlp/tweet\_eval}}
Statistics on the datasets used in the experiments are shown in Table~\ref{tab:clustering_dataset}.
Overall, there are between 1,000 and 10,000 instances.

\begin{table}[h!]
\small
\centering
\begin{tabular}{cccc} \\ 
\toprule
Dataset & \#Labels & \#Val. & \#Test \\
\midrule
C-STS & - & 2,840 & 4,732 \\
\midrule
Amazon-C & 31 & 5,000 & 5,000 \\
Amazon-R & 5 & 5,000 & 5,000 \\
ScienceQA & 25 & 4,241 & 4,241\\
Tweet Emotion & 4 & 374 & 1,421 \\
\bottomrule
\end{tabular}
\vspace{-5pt}
\caption{Statistics on datasets used in the experiments}
\label{tab:clustering_dataset}
\end{table}

\section{Model links used in PonTE}
\label{sec:specific_models}
We used six LLMs as the model in the Hugging Face Hub~\cite{wolf-etal-2020-transformers} for PonTE.
Specifically, we used the base\footnote{\href{https://huggingface.co/mistralai/Mistral-7B-v0.3}{mistralai/Mistral-7B-v0.3}} and instruction-tuned\footnote{\href{https://huggingface.co/mistralai/Mistral-7B-Instruct-v0.3}{mistralai/Mistral-7B-Instruct-v0.3}} models of Mistral-7B~\cite{jiang2023mistral7b}, the base\footnote{\href{https://huggingface.co/meta-llama/Meta-Llama-3-8B}{meta-llama/Meta-Llama-3-8B}} and instruction-tuned\footnote{\href{https://huggingface.co/meta-llama/Meta-Llama-3-8B-Instruct}{meta-llama/Meta-Llama-3-8B-Instruct}} models of Llama-3-8B~\cite{touvron2023llama}, and the base\footnote{\href{https://huggingface.co/meta-llama/Meta-Llama-3-70B}{meta-llama/Meta-Llama-3-70B}} and instruction-tuned\footnote{\href{https://huggingface.co/meta-llama/Meta-Llama-3-70B-Instruct}{meta-llama/Meta-Llama-3-70B-Instruct}} models of Llama-3-70B.

\section{Explanation of Previous Methods}
\label{sec:additional_explanation_previous_methods}
We describe previous methods that we compared to PonTE in our experiments.

\paragraph{Universal Text Embeddings}
Universal text embedding refers to a single, versatile vector representation for a single text, designed to place semantic similar texts closely together in a semantic vector space.
One of the representative methods is SimCSE~\cite{gao-etal-2021-simcse}, which includes unsupervised and supervised methods using contrastive learning.
The unsupervised method involved training on its own instance with added noise owing to dropout masks as a positive example and other instances as negative examples, and the supervised method involved fine-tuning using text pair data, such as NLI.
We used unsupervised\footnote{\href{https://huggingface.co/princeton-nlp/unsup-simcse-roberta-large}{princeton-nlp/unsup-simcse-roberta-large}} and supervised\footnote{\href{https://huggingface.co/princeton-nlp/sup-simcse-roberta-large}{princeton-nlp/sup-simcse-roberta-large}} SimCSE.
In C-STS, the text and condition are concatenated and entered into the model whereas in text clustering, only the text is entered.

PromptEOL~\cite{jiang2023scaling} used prompts that can compress textual information into a single word.
PromptEOL utilizes `\textit{This sentence: ``{\rm \{sentence\}}'' means in one word: ``}', where the target sentence is entered in `\{sentence\}', as the prompt template.
This draws inspiration from PromptBERT~\cite{jiang-etal-2022-promptbert}, which is a prompt-based method that generates universal text embeddings by fine-tuning masked language models, such as BERT~\cite{devlin-etal-2019-bert}.
In the experiment, the same causal LLMs as PonTE were used to measure performance when no condition was specified.

\paragraph{Conditional Text Embeddings}
Conditional text embedding is a textual representation that allows for precise control of the destination of text embedding by specifying conditions.
For example, the instruction fine-tuned embedding methods, which are a type of conditional text embedding method, generate task-specific text embeddings based on different instructions for each task.
In our experiments, we used two methods as a baseline.
The first method was GTE~\cite{li2023generaltextembeddingsmultistage},\footnote{\href{https://huggingface.co/Alibaba-NLP/gte-Qwen2-7B-instruct}{Alibaba-NLP/gte-Qwen2-7B-instruct}} which is based on the instruction-tuned model of Qwen2-7B. 
This method utilizes unsupervised contrastive learning on massive text pairs mined from the Web, followed by supervised contrastive learning on various manually curated texts.
The other method was E5~\cite{wang2024textembeddingsweaklysupervisedcontrastive,wang2024improvingtextembeddingslarge},\footnote{\href{https://huggingface.co/intfloat/e5-mistral-7b-instruct}{intfloat/e5-mistral-7b-instruct}} which is based on the instruction-tuned model of Mistral-7B. 

Several methods of fine-tuning using C-STS datasets have also been proposed.
\citet{deshpande-etal-2023-c} trained bi-encoder or tri-encoder architectures based on RoBERTa~\cite{liu2019roberta}, DiffCSE~\cite{chuang-etal-2022-diffcse}, and SimCSE~\cite{gao-etal-2021-simcse} on the C-STS dataset.
By inputting the target and conditional texts into these models, conditional text embeddings can be obtained.
\citet{yoo2024hyper} proposed Hyper-CL~\cite{yoo2024hyper}, which enhanced the interaction between the target text and conditional text of the tri-encoder method.
Since these models leverage in-domain data, these models are outside our scope, but their scores are provided in Table~\ref{tab:experiments_csts_supervised} for reference to score upper bounds.
It can be seen that the score is high for Bi-Encoder but there is little difference in performance between the Tri-Encoder methods and PonTE using Llama-3-7B-Inst in Table~\ref{tab:experiments_csts}.

\begin{table}[t!]
\vspace{-10pt}
\small
\centering
\begin{tabular}{l cc} \\ 
\toprule
Method & $\rho$ & $r$ \\ 
\midrule
$^\dag$Bi-Encoder$_{\textrm{\footnotesize sup-SimCSE-large}}$ & \textbf{47.5} & \textbf{47.6} \\
$^\dag$Tri-Encoder$_{\textrm{\footnotesize sup-SimCSE-large}}$ & 35.3 & 35.6 \\
$^\ddag$Tri-Encoder$_{\textrm{\footnotesize sup-SimCSE-large+hyper-cl}}$ & 39.6 & 40.0 \\ 
\bottomrule
\end{tabular}
\vspace{-5pt}
\caption{Experimental results on C-STS using previous supervised conditional text embedding methods. 
$^\dag$ and $^\ddag$ denote the results from \citet{deshpande-etal-2023-c} and \citet{yoo2024hyper}, respectively.}
\vspace{-15pt}
\label{tab:experiments_csts_supervised}
\end{table}

\begin{table*}[t!]
\vspace{-15pt}
\small
\centering
\begin{tabular}{c@{\ \ }l cc} \\ 
\toprule
\multicolumn{2}{l}{Prompt template} & $\rho$ & $p$ \\ 
\midrule
(1) & This text: ``\{text\}'' means \textbf{in terms of \{condition\}}: `` & 18.8 & 17.1 \\ 
(2) & This text: ``\{text\}'' means \textbf{with respect to \{condition\}}: `` & 18.0 & 17.0 \\ 
(3) & This text: ``\{text\}'' means in one word \textbf{in terms of \{condition\}}: `` & 28.2 & 25.2 \\ 
(4) & This text: ``\{text\}'' means in one word \textbf{with respect to \{condition\}}: `` & 25.1 & 21.7 \\ 
(5) & This text: ``\{text\}'' means \textbf{in terms of \{condition\}} in one word: `` & 28.1 & 24.7 \\ 
(6) & This text: ``\{text\}'' means \textbf{with respect to \{condition\}} in one word: `` & 25.4 & 22.3 \\ \midrule
(7) & Express this text ``\{text\}'' \textbf{in terms of \{condition\}}: `` & 19.8 & 18.2 \\ 
(8) & Express this text ``\{text\}'' \textbf{with respect to \{condition\}}: `` & 19.1 & 17.6 \\ 
(9) & Express this text ``\{text\}'' in one word \textbf{in terms of \{condition\}}: `` & \textbf{37.3} & \textbf{34.8} \\ 
(10) & Express this text ``\{text\}'' in one word \textbf{with respect to \{condition\}}: `` & 36.4 & 33.9 \\ 
(11) & Express this text ``\{text\}'' \textbf{in terms of \{condition\}} in one word: `` & 33.1 & 30.4 \\ 
(12) & Express this text ``\{text\}'' \textbf{with respect to \{condition\}} in one word: `` & 30.4 & 27.9 \\ \bottomrule
\end{tabular}
\vspace{-5pt}
\caption{Prompt templates and the Spearman ($\rho$) and Pearson ($p$) correlation coefficients of PonTE$_{\textrm{\footnotesize Llama-3-8B-Inst}}$ on the validation set in the C-STS experiment.}
\label{tab:prompt_template}
\end{table*}

\section{Prompt Template Search on C-STS}
\label{sec:csts_prompt_template_search}
Prompt templates and the Spearman and Pearson correlation coefficients of PonTE$_{\textrm{\footnotesize Llama-3-8B-Inst}}$ on the validation set in the C-STS experiment are listed in Table~\ref{tab:prompt_template}.
Prompts from (1) to (6) incorporate additional conditional text into the prompt template proposed in PromptEOL, whereas prompts (7) to (12) introduce new prompt templates that leverage the instruction-tuned model.

\begin{table}[t!]
\vspace{-15pt}
\small
\centering
\begin{tabular}{l ccc} \\ 
\toprule
Method & PT & $\rho$ & $r$ \\ 
\midrule \midrule
\multicolumn{4}{c}{\textit{Supervised method}} \\ 
\midrule
sup-SimCSE$_{\textrm{\footnotesize large}}$ & - & 0.2 & 0.7 \\ 
\midrule
GTE$_{\textrm{\footnotesize Qwen2-7B-Inst}}$ & - & 27.6 & 28.1 \\
E5$_{\textrm{\footnotesize Mistral-7B-Inst}}$ & - & \textbf{32.3} & \textbf{32.3} \\
\midrule \midrule

\multicolumn{4}{c}{\textit{Unsupervised method}} \\ 
\midrule
unsup-SimCSE$_{\textrm{\footnotesize large}}$ & - & -1.8 & -1.7 \\ 
\midrule
PonTE$_{\textrm{\footnotesize Mistral-7B}}$ & (6) & 21.0 & 20.2 \\
PonTE$_{\textrm{\footnotesize Mistral-7B-Inst}}$ & (6) & 31.8 & 30.3 \\
PonTE$_{\textrm{\footnotesize Llama-3-8B}}$ & (10) & 17.9 & 16.4 \\
PonTE$_{\textrm{\footnotesize Llama-3-8B-Inst}}$ & (9) & \textbf{37.3} & \textbf{34.8} \\ 
PonTE$_{\textrm{\footnotesize Llama-3-70B}}$ & (11) & 16.3 & 15.4 \\
PonTE$_{\textrm{\footnotesize Llama-3-70B-Inst}}$ & (9) & 35.8 & 32.4 \\ 
\bottomrule
\end{tabular}
\vspace{-5pt}
\caption{Experimental results on the validation set for C-STS. 
The prompt template (PT) used for each method corresponds to the bracketed numbers in Table~\ref{tab:prompt_template}.}
\label{tab:csts_validation_set}
\vspace{-10pt}
\end{table}

Based on the experimental results, the scores were higher when ``in one word'' was inserted, compared with when it was not.
This demonstrates the effectiveness of explicit one-word prompts.
In addition, prompt templates starting with ``Express'' from (7) to (12) yielded higher overall scores compared with those starting with ``This text'' from (1) to (6).
This is attributed to the fact that using instructions, such as ``Express XXX,'' helps the instruction-tuned model to better understand the prompt’s intent.
The differences in the use of ``in terms of'' and ``with respect to'' were minimal.

\section{Results on Validation Set for C-STS}
\label{sec:csts_validation_set}
Table~\ref{tab:csts_validation_set} lists the Spearman and Pearson correlation coefficients on the validation set in the C-STS.
The prompt template corresponding to Table~\ref{tab:prompt_template}, which had the highest Spearman correlation coefficient in PonTE using each model, is also shown.
The trends of the scores are roughly the same as in Table~\ref{tab:experiments_csts}.
The prompt template used in PonTE differs among the models, but they all share the use of ``in one word.''
The prompt template with the highest Spearman correlation coefficient in the validation set for C-STS is used as the prompt template for each model in the text clustering.

\begin{table}[t!]
\vspace{-15pt}
\small
\centering
\begin{tabular}{c@{\ \ }c@{\ \ }l@{\ \ }c} \\ 
\toprule
Dataset & \multicolumn{2}{l}{Conditional text} & Score \\ 
\midrule
\multirow{6}{*}{Amazon-C}
& (a) & the category & 22.8 \\ 
& (b) & the product category & 26.5 \\ 
& (c) & the category name & 21.6 \\ 
& (d) & the product category name & 29.4 \\ 
& (e) & the name of the category & 22.0 \\ 
& (f) & the name of the product category & \textbf{30.5} \\ 

\midrule

\multirow{6}{*}{Amazon-R}
& (g) & the rating & 30.4 \\ 
& (h) & the star & 26.2 \\ 
& (i) & the star rating & \textbf{34.1} \\ 
& (j) & the five-level rating & 21.3 \\ 
& (k) & the five-level star rating & 29.3 \\ 
& (l) & the emotion & 33.4 \\

\midrule

\multirow{4}{*}{SciQA}
& (m) & the category & 70.3 \\
& (n) & the question category & 74.2 \\ 
& (o) & the name of the category & 74.1 \\ 
& (p) & the name of the question category & \textbf{75.4} \\ 

\midrule

\multirow{2}{*}{Tweet}
& (q) & the emotion & \textbf{45.9} \\
\multirow{2}{*}{Emotion}
& (r) & the feeling & 44.7 \\ 
& (s) & the sentiment & 43.6 \\ 

\bottomrule
\end{tabular}
\vspace{-5pt}
\caption{Conditional texts in prompt templates and V-measure scores of PonTE$_{\textrm{\footnotesize Llama-3-8B-Inst}}$ on the validation set in the text clustering experiments}
\label{tab:condition_text}
\end{table}

\begin{table*}[t!]
\vspace{-5pt}
\small
\centering
\begin{tabular}{l cccc}
\toprule
\multirow{2}{*}{Method} & \multicolumn{2}{@{}c}{Amazon} & \multirow{2}{*}{SciQA} & Tweet \\
& -C & -R & & Emotion \\
\midrule \midrule

\multicolumn{5}{c}{\textit{Supervised method}} \\ 
\midrule
sup-SimCSE$_{\textrm{\footnotesize large}}$ & 18.1 & 24.4 & 65.6 & 24.1 \\
\midrule

GTE$_{\textrm{\footnotesize Qwen2-7B-Inst}}$ & \textbf{37.4} & 37.0 & \textbf{72.1} & \textbf{41.6} \\
E5$_{\textrm{\footnotesize Mistral-7B-Inst}}$ & 35.6 & \textbf{37.6} & 70.3 & 38.5 \\

\midrule \midrule

\multicolumn{5}{c}{\textit{Unsupervised method}} \\ 
\midrule

unsup-SimCSE$_{\textrm{\footnotesize large}}$ & 16.1 & 4.0 & 60.9 & 8.3 \\
\midrule

PromptEOL$_{\textrm{\footnotesize Mistral-7B}}$ & 8.1 & 27.5 & 64.3 & 8.3 \\
PromptEOL$_{\textrm{\footnotesize Mistral-7B-Inst}}$ & 6.0 & 28.3 & 59.7 & 21.2 \\
PromptEOL$_{\textrm{\footnotesize Llama-3-8B}}$ & 9.2 & 21.0 & 68.9 & 6.5 \\
PromptEOL$_{\textrm{\footnotesize Llama-3-8B-Inst}}$ & 8.7 & 31.9 & 66.1 & 27.0 \\
\midrule

PonTE$_{\textrm{\footnotesize Mistral-7B}}$ & 28.1 (d) & 27.8 (g) & 73.1 (n) & 27.0 (r) \\
PonTE$_{\textrm{\footnotesize Mistral-7B-Inst}}$ & 25.3 (d) & 32.1 (g) & 69.7 (n) & 43.4 (s) \\
PonTE$_{\textrm{\footnotesize Llama-3-8B}}$ & \textbf{30.8} (f) & 21.9 (l) & 74.4 (m) & 23.2 (q) \\
PonTE$_{\textrm{\footnotesize Llama-3-8B-Inst}}$ & 30.0 (f) & \textbf{34.0} (i) & \textbf{75.4} (p) & \textbf{45.9} (q) \\
\bottomrule
\end{tabular}
\vspace{-5pt}
\caption{Experimental results on the validation set for text clustering.
The score represents the V-measure value.
The bracketed letters are conditional text corresponding to Table~\ref{tab:condition_text}, which shows the highest performance in the validation set.}
\label{tab:text_clustering_validation}
\vspace{-10pt}
\end{table*}

\section{Conditional Text Search for Prompt using Text Clustering}
\label{sec:clustering_condition_search}
Conditional texts in the prompt template and V-measure scores of PonTE$_{\textrm{\footnotesize Llama-3-8B-Inst}}$ on the validation set in the text clustering experiments for each dataset are listed in Table~\ref{tab:condition_text}.
On the Amazon-C and SciQA, adding ``product'' or ``name'' enhances performance.
Additionally, on the Amazon-R, using ``star rating'' instead of just ``star'' or ``rating'' resulted in higher scores.
This trend is attributed to the increased specificity of the conditional text, enabling the LLM to better comprehend the intent of the prompt.

\section{Results on Validation Set for Text Clustering}
\label{sec:text_clustering_validation}

Table~\ref{tab:text_clustering_validation} shows the V-measure scores on the validation set for text clustering for each method.
It also displays the conditional text with the highest score for each model in PonTE, corresponding to Table~\ref{tab:condition_text}.
The trend in scores is much the same as for the test set.
Similar to the results for the prompt templates used in each model in Table~\ref{tab:csts_validation_set}, the conditional text used in each model tends to be quite different.
We need to search for the best conditional text for each model.
\end{document}